\documentclass[letterpaper, 10pt, conference]{ieeeconf}  % Comment this line out if you need a4paper
\IEEEoverridecommandlockouts                              % This command is only needed if 
\usepackage{cite}
\usepackage{graphicx}
\usepackage{amsmath}
\usepackage{amssymb}
\usepackage{algorithm}
\usepackage[noend]{algorithmic}
\usepackage{flushend}
\usepackage[T1]{fontenc}
\usepackage{aecompl}

\newcommand{\Acronym}[1]{\ensuremath{{{\texttt{#1}}}}}
\newcommand{\Symbol}[1]{\ensuremath{\mathcal{#1}}}
\newcommand{\Function}[1]{\ensuremath{{\textsc{#1}}}}

\newcommand{\Var}[1]{\ensuremath{{{\mathrm{#1}}}}}

\newcommand{\Null}{\Acronym{null}}

\newcommand{\Pair}[1]{\ensuremath{{\langle #1 \rangle}}}
\newcommand{\World}{\Symbol{W}}
\newcommand{\Obstacles}{\Symbol{O}}
\newcommand{\Obstacle}{\Symbol{O}}
\newcommand{\Goals}{\Symbol{G}}
\newcommand{\Goal}{\Symbol{G}}

\newcommand{\Robot}{\Symbol{R}}
\newcommand{\Shape}{\Symbol{P}}

\newcommand{\StateSpace}{\Symbol{S}}
\newcommand{\ActionSpace}{\Symbol{U}}
\newcommand{\MotionEqs}{\ensuremath{f}}
\newcommand{\Simulate}{\Function{simulate}}
\newcommand{\Traj}{\ensuremath{\zeta}}

\newcommand{\Tree}{\Symbol{T}}

\newcommand{\TreeNode}{\ensuremath{\eta}}

\newcommand{\Group}{\ensuremath{\Gamma}}

\newcommand{\CostMatrix}{\ensuremath{\Delta}}

\DeclareMathOperator*{\argmin}{argmin}

\title{Multi-Goal Motion Memory}

\author{Yuanjie Lu$^1$, Dibyendu Das$^1$, Erion Plaku$^2$, and Xuesu Xiao$^1$% <-this % stops a space
\thanks{$^1$Department of Computer Science, George Mason University, Fairfax, VA 22030, USA.}
\thanks{$^2$National Science Foundation, Alexandria,
VA 22314, USA}%
\thanks{The work by E. Plaku is supported by (while serving at) the National Science Foundation. Any opinion, findings, and conclusions or recommendations expressed in this material are those of the authors and do not necessarily reflect the views of the National Science Foundation.}
}

\begin{document}

\maketitle

\begin{abstract} 

Autonomous mobile robots (e.g., warehouse logistics robots) often need to traverse complex, obstacle-rich, and changing environments to reach multiple fixed goals (e.g., warehouse shelves). 
Traditional motion planners need to calculate the entire multi-goal path from scratch in response to changes in the environment, which result in a large consumption of computing resources. 
This process is not only time-consuming but also may not meet real-time requirements in application scenarios that require rapid response to environmental changes. 
In this paper, we provide a novel Multi-Goal Motion Memory technique that allows robots to use previous planning experiences to accelerate future multi-goal planning in changing environments. Specifically, our technique predicts collision-free and dynamically-feasible trajectories and distances between goal pairs to guide the sampling process to build a roadmap, to inform a Traveling Salesman Problem (TSP) solver to compute a tour, and to efficiently produce motion plans. 
Experiments conducted with a vehicle and a snake-like robot in obstacle-rich environments show that the proposed Motion Memory technique can substantially accelerate planning speed by up to 90\%. Furthermore, the solution quality is comparable to state-of-the-art algorithms and even better in some environments.

\end{abstract}

%\begin{IEEEkeywords}
%xxx
%\end{IEEEkeywords}

\section{Introduction}
\label{sec:Intro}

Robots used in applications such as logistics, inspection, surveillance, and transportation often have to reach multiple fixed locations to carry out assigned tasks\cite{book:MP,book:LaValle}. Their workspaces, such as warehouses, factories, and power plants, are usually unstructured and contain numerous obstacles, requiring the robots to navigate around them and pass through several narrow passages. Therefore, working in these settings give rise to challenging multi-goal motion-planning problems, where robots need to quickly plan their motions to reach the goal locations while avoiding obstacles. 

In a multi-goal motion-planning setting, a robot has to reason at a high level to determine the order in which to reach the goals, bearing similarities with Traveling Salesman Problems (TSPs). However, in this case, the costs and trajectories of moving from one location to another are not known beforehand, as using straight-line paths and Euclidean distances as a proxy does not account for the obstacles or robot dynamics (i.e., limiting the direction of motion, velocity, turn radius, acceleration, etc.). Therefore, multi-goal motion planning requires solving numerous single-goal motion-planning problems. Another challenge is that high-level reasoning and motion planning cannot be decoupled. Computing a goal ordering and then running a motion planner to reach the goals in succession will likely lead to suboptimal solutions, since the goal ordering has not necessarily accounted for the obstacles and robot dynamics. Thus, the vast spaces of feasible motion plans and goal orderings must be simultaneously explored. 

To make matters worse, while classical motion planners can solve such multi-goal motion planning problems in conjunction with TSP solvers~\cite{mcmahon2021dynamic} and satisfy the physical constraints imposed by the robot dynamics, they need to start planning from scratch each time the environment changes, regardless of how small the change is (e.g., some items are moved on a factory floor). Such repetitive planning not only leads to a waste of computation but also limits the robot's ability to respond quickly when facing environmental changes, which are not uncommon in real-world applications. 

\begin{figure}[t]
\centering
\includegraphics[width=0.9\columnwidth]{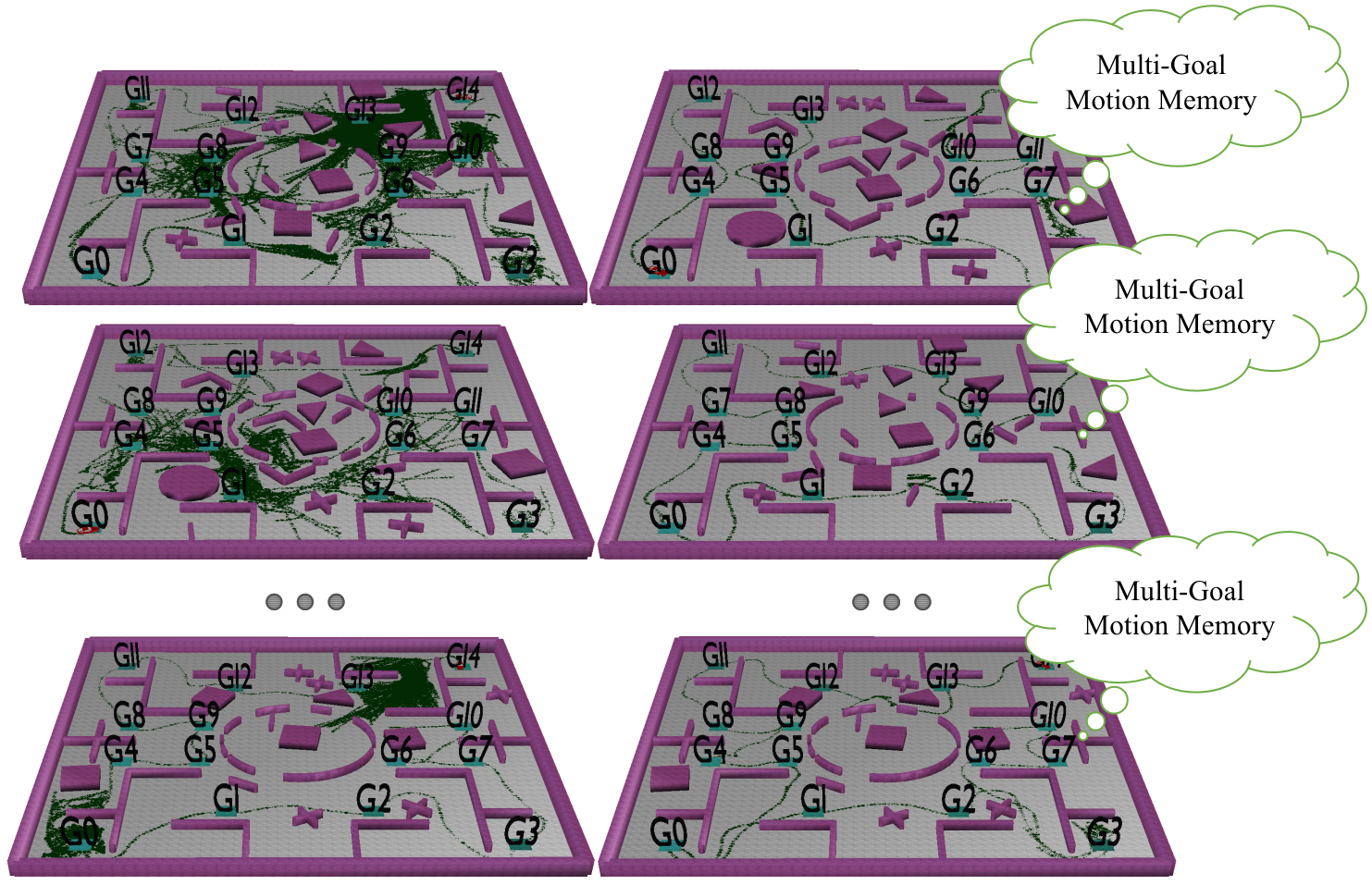}
\caption{In multi-goal motion planning, the robot must visit each goal ($G_i, \forall i$) while avoiding collisions. Classical multi-goal motion planners require extensive computation when facing a new planning problem (shown in green), which can be significantly reduced by Multi-Goal Motion Memory. }
\label{fig:Main}
\end{figure}

To address these challenges, we present a new technique called Multi-Goal Motion Memory, which incorporates past planning experiences into a planner and uses this prior knowledge to accelerate new planning tasks (Fig.~\ref{fig:Main}). Specifically, leveraging historical planning experiences, Multi-Goal Motion Memory quickly generates the best trajectory with dynamics and the predicted distance between each goal pair to guide the sampling process for building a new motion map for the new environment. Then, utilizing the generated goal-to-goal distances, a TSP solver efficiently computes a low-cost tour that determines the optimal order to visit each goal. After determining the order of accessing the goals, the predicted best trajectory is used to guide the motion-tree expansion in a motion planner, 
% When attempting to expand a node in the motion tree toward the next goal in the tour, several states are sampled near the predicted trajectory, and priority is given to those ?states that are associated with low-cost predictions. Such a priority 
which allows the motion tree to more effectively explore low-cost areas, considerably reducing the motion-planning runtime.
Experiments are conducted in simulation using a vehicle and snake-like robot operating in unstructured, obstacle-rich environments with an increasing number of goals. The results show that Multi-Goal Motion Memory can significantly accelerate planning speed by up to 90\%. Moreover, the plan quality in terms of solution distance remains comparable to state-of-the-art algorithms and is even better in many environments.

\section{Related Work}
\label{sec:RelatedWork}

We review related work in multi-goal motion planning and machine learning for motion planning. 

\subsection{Multi-Goal Motion Planning}

Multi-goal motion planning is the problem of finding a collision-free and dynamically-feasible trajectory to visit a set of goals. Researchers have addressed different aspects of this problem. At a high level, which does not account for the obstacles or the vehicle dynamics, the focus has been on computing optimal tours based on goal-pair distances via TSP solvers~\cite{gutin2006traveling,applegate2011traveling,helsgaun2017extension, FaiglTSP,bogyrbayeva2023deep}. When considering obstacles but not robot dynamics, general motion planners often construct a roadmap based on PRM~\cite{PRM}. Researchers have used TSP solvers on a cost matrix based on the shortest path distance to find paths to solve multi-goal motion planning \cite{danner2000randomized,saha2006planning,englot2012sampling}. 
TSP solvers have also been combined with controllers for Dubins vehicles \cite{jang2017optimal,janovs2022randomized}. Sampling-based motion planning~\cite{edelkamp2014multi,rashidian2014motion,edelkamp2018integrating} has often been used for solving multi-goal problems with robot dynamics. 

Despite the potential similarity among planning problems in the workspace, these classical multi-goal motion planners re-plan from scratch in response to any environmental changes. 
Multi-Goal Motion Memory is designed to address such a limitation by leveraging previous experiences to more efficiently guide the current motion-tree expansion. %To be specific, we build upon a classical multi-goal motion planner, DROMOS~\cite{plaku2018multi}, which uses a sampled roadmap, a TSP solver, and a single-goal motion planner to plan from scratch when given a new planning problem. 
% Another limitation of DROMOS is that it requires sufficient configurations to ensure that every goals can be connected to. Once one of the paths cannot be found, the entire process will be considered failed, wasting significant time and resources.

\subsection{Machine Learning for Motion Planning}

In recent years, researchers have started to use machine learning to solve single-goal motion planning~\cite{mcmahon2022survey, xiao2022motion, wang2020neural, li2018neural,ichter2020learned, rajamaki2017augmenting, wang2021agile, xiao2021agile, xiao2021toward}. Reinforcement and imitation learning~\cite{xiao2022motion} has been used for motion planning in complex environments, involving highly-cluttered obstacles~\cite{wang2021agile, xiao2021agile, xiao2021toward, xiao2022autonomous, xiao2022appl, xiao2020appld, wang2021appli, wang2021apple, xu2023benchmarking, xu2021machine}, dynamic decision-making agents~\cite{everett2018motion, xiao2022learning, Karnan2022a, Nguyen2023, mirsky2021conflict}, and off-road conditions~\cite{xiao2021learning, karnan2022vi, sikand2022visual, datar2023toward}. The work in \cite{qureshi2019motion} presents a neural-network-based planning algorithm to encode the given workspaces from a point cloud measurement and generate collision-free paths from the start to the goal. The work in \cite{pan2015efficient} combines machine learning and geometric approximation to build a new collision-free configuration space, using parallel $k$-nearest neighbors and parallel collision detection algorithms to accelerate motion planning. In terms of sampling-based motion planners, the work in \cite{zhang2018learning} presents a policy-search-based method to learn implicit sampling distributions for different environments, making it possible to generate better distributions and reduce cost.
Machine learning has also been used to guide the motion-tree expansion by first predicting if a given sample is in a collision and then estimating if it is a promising sample integrated into RRT~\cite{arslan2015machine}.
The work in \cite{buiimproving} proposed the idea of training a model to predict the runtime on single-goal problems, which helps to facilitate the construction of cost matrices for multi-goal problems.

The most relevant work is our previous approach~\cite{lu2023leveraging} to improve the planning efficiency of DROMOS~\cite{plaku2018multi}. Given the same environment, the work in \cite{lu2023leveraging} learns to predict goal-to-goal distances for any arbitrary goal pair using a large number of collision-free configurations and motion plans by DROMOS. Our Multi-Goal Motion Memory is based on the observation that in many real-world robotics applications, the goal locations remain constant while the environment keeps changing, necessitating a way to quickly plan in response to different obstacle configurations. 

% To overcome these limitations, we leverage prior single-goal motion memory work~\cite{das2023motion} for our multi-goal motion planning problems. The work is based on planning experience, learning, and summarizing motion planning with dynamics in different environments in the past to solve single-goal planning problems. We can use these advantages to solve any pair of goals in multi-goal problems. 

\section{Preliminaries}
\label{sec:preliminaries}
This section defines the problem of multi-goal motion planning with robot dynamics. It also presents a general pipeline for many classical sampling-based multi-goal motion planners and sets the stage for potential data-driven improvement. The section concludes with a description of the single-start-goal motion memory technique, which our Multi-Goal Motion Memory generalizes. 

\subsection{Problem Definition}
\label{sec:Problem}
A robot operates in a world $\World$ containing obstacles $\Obstacles=\{\Obstacle_1, \ldots, \Obstacle_m\}$ and  goal regions $\Goals=\{\Goal_1, \ldots, \Goal_n\}$, as shown in Fig.~\ref{fig:Main}. The robot is modeled as $\Robot=\Pair{\Shape, \StateSpace, \ActionSpace, \MotionEqs}$ in terms of its geometric shape $\Shape$, state space $\StateSpace$, action space $\ActionSpace$, and dynamics $\MotionEqs$ expressed as a set of differential equations: 
\begin{equation}
\dot{s} = f(s, a), s \in \StateSpace, a \in \ActionSpace.\nonumber
\end{equation}
A vehicle and snake-like robot are shown in Fig.~\ref{fig::env} as examples. The vehicle robot is defined as $s = (x, y, \theta, \psi, v)$ in terms of the position $(x, y)$, orientation $\theta$, steering angle $\psi $ ($|\psi| \leq 1.5\text{rad}$), and speed $v$ ($|v| \leq 2.25m/s$). The vehicle is controlled by setting the acceleration $a_\Var{acc}$ ($|a_\Var{acc}| \leq 1m/s^2$) and the steering turning rate $a_\omega$ ($|\omega| \leq 2.7\text{rad}/s$). The motion equations are defined as 
\begin{equation}
\dot{x} = v\cos{(\theta)}\cos{(\psi)}, \dot{y} = v\sin{(\theta)}\cos{(\psi)}, \nonumber
\end{equation}
\begin{equation}
\dot{\theta} = v\sin{(\psi)} / L, \dot{v} = a_{acc}, \dot{\psi} = a_w, \nonumber
\end{equation}
where $L$ is the distance from the back to the front wheels. The snake-like robot, modeled as a car pulling trailers, is defined as $s = (x, y, \theta, \psi, v, \theta_1, \dots, \theta_N)$, and $f$ includes 
\begin{equation}
\dot{\theta_i} = \frac{v}{H}(\sin{\theta_{i-1}} - \sin{\theta_0)}\prod^{i-1}_{j-1}\cos{(\theta_{j-1} -\theta_j)}, \nonumber
\end{equation}
where $\theta_0 = \theta$ and the hitch distance $H$ is set to a small value so that the robot resembles a snake.

A state $s \in \StateSpace$ reaches a goal $\Goal_i$ if the position defined by $s$ is inside $\Goal_i$.
The state $s$ is in collision if the robot overlaps with an obstacle when placed according to the position and orientation defined by $s$. 

Applying a control action $a \in \ActionSpace$ to a state $s \in \StateSpace$ yields a new state $s_\Var{new} \in \StateSpace$, which is obtained by numerically integrating the motion equations $\MotionEqs$ for one time step $dt$, i.e.,
\begin{equation}
s_{new} \gets \Simulate(s, a, f, dt). \nonumber
\end{equation}
A dynamically-feasible trajectory $\Traj: \{0, \ldots, \ell\} \rightarrow \StateSpace$ is obtained by applying a sequence of actions $\Pair{a_0, \ldots, a_{\ell-1}}$ in succession, where $\Traj_0 \leftarrow s$ and $\forall j \in \{1, \ldots, \ell\}$:
\begin{equation}
\Traj_{j} \gets \Simulate(\Traj_{j-1}, a_{j-1}, f, dt).
\label{eqn:simulate}
\end{equation}
 The multi-goal motion-planning problem is defined as follows: Given a world $\World$ containing obstacles  $\Obstacles=\{\Obstacle_1, \ldots, \Obstacle_m\}$ and goals $\Goals=\{\Goal_1, \ldots, \Goal_n\}$, a  robot model $\Robot=\Pair{\Shape, \StateSpace, \ActionSpace, \MotionEqs}$, and an initial state $s_\Var{init} \in \StateSpace$, compute a collision-free and dynamically-feasible trajectory $\Traj : \{0, \ldots, \ell\} \rightarrow \StateSpace$ that starts at $s_\Var{init}$ and reaches each goal. The objective of this work is to reduce the planning time.

\subsection{Classical Sampling-Based Multi-Goal Motion Planning}
Classical sampling-based multi-goal motion planning is primarily divided into three parts (Algorithm~\ref{alg:multi-goal-motion-planner}). In line 1, a roadmap is constructed and represented as an undirected weighted graph by sampling collision-free states. The roadmap retains information about the coordinates of each state and the cost of each edge. Subsequently, through Dijkstra's algorithm, we can obtain the path from each collision-free state to any goal and its cost. In line 2, the cost from the robot position to each goal and from each goal to every other goal is sent to a TSP solver, which then computes a tour visiting all the goals. In line 3, a motion planner will plan the path to reach all the goals based on the order specified by the tour. 
% During the extension process, each status update will perform a TSP, and then continuously obtain the path, and finally reach each target point.
Our Multi-Goal Motion Memory aims at improving all three parts in a data-driven manner. 

\begin{algorithm}[!t]
\caption{General Multi-Goal Motion Planner}
\label{alg:multi-goal-motion-planner}

\textbf{Input}: world $\mathcal{W}$; obstacles $\mathcal{O}$; goals $\mathcal{G}$; robot model $\mathcal{R}$; initial state $s_\Var{init}$; $\Function{TSPsolver}$ 
\\ 
\textbf{Output}:  collision-free and dynamically-feasible trajectory that starts at $s_\Var{init}$ and reaches each goal in $\Goals$
\vspace*{1mm}
\hrule

\begin{algorithmic}[1]
\STATE $\mathcal{RM} \gets \Function{CreateRoadMap}(\mathcal{W}, \mathcal{O}, \mathcal{G}, s_\Var{init})$
\STATE $\Var{tour} \gets \Function{TSPsolver}(\mathcal{RM}, \mathcal{G}, \mathcal{R}, s_\Var{init})$
\STATE $p \gets \Function{PlanWithMotionTree}(\Var{tour}, \mathcal{R})$
\end{algorithmic}
\end{algorithm}

\subsection{Single-Start-Goal Motion Memory}

Motion Memory is a recently proposed approach~\cite{das2023motion} with an experience augmentation technique and a representation learning method that enable robots to reflect on prior planning experiences for efficient future planning for a single start-goal pair~\cite{das2023motion}. 
Given a set of existing motion planning experience for a fixed start and goal, i.e., a mapping from an obstacle configuration to a motion plan, motion memory first produces a variety of similar obstacle configurations where the same motion plan would still be optimal. 
Subsequently, this augmented dataset is utilized to learn an efficient latent representation space by minimizing a triplet loss function. 
As a result, the representation space is organized into several latent clusters, with each cluster representing a distinct motion plan from past experiences. When a new planning problem arises, it is projected into this latent space. The motion plan associated with the nearest cluster centroid is then considered the most appropriate approximation to the actual motion planning solution for the new problem. 

Our Multi-Goal Motion Memory is a generalization of the specific single-goal case. By generalizing motion memory to consider different start-goal pairs, it can be efficiently used to solve multi-goal motion planning problems by accelerating the solution time of multiple components (Algorithm~\ref{alg:multi-goal-motion-planner}). 
\section{Methodology}
\label{sec::method}

Motivated by many real-world multi-goal motion planning problems with a fixed goal set (e.g., stations or targets the robot needs to visit) but changing obstacles, Multi-Goal Motion Memory aims to leverage past planning experiences to accelerate future planning given new environments. In principle, Multi-Goal Motion Memory can be used in conjunction with any multi-goal motion planner. We first introduce Multi-Goal Motion Memory in a planner-agnostic manner, and then we integrate Multi-Goal Motion Memory into sampling-based multi-goal motion planners based on TSP solvers following Algorithm~\ref{alg:multi-goal-motion-planner} at different stages.  

\subsection{Multi-Goal Motion Memory}

A particular environment's State Space (S-space, $S$) can be decomposed into reachable ($S_{free}$) and unreachable ($S_{obst}$ determined by $\Obstacles$) states, considering various constraints like obstacles, nonholonomic constraints, and velocity limits. 
Given a a start state ($S_s\in S$) and a goal state ($S_g\in S$), a motion plan, denoted as \(P=\{a_i | 1\leq i \leq T\}\), where $a_i\in\mathcal{U}$, drives the robot through $S_{free}$ by optimizing a cost function and observing robot dynamics. 

The previous single-start-goal motion memory assumes all planning problems have the same start and goal pair. Therefore, solving a motion planning problem is to produce a motion planning solution based on the S-space alone. A dataset of motion planning problems and solutions for a fixed  start-goal pair becomes $\mathcal{D}^S=\{d_i\}_{i=1}^N=\{S_i, P_i\}_{i=1}^N$ and the motion planner becomes a function $f(\cdot)$ that maps from S-spaces to motion plans, i.e., $P=f(S)$. Single-start-goal motion memory also utilizes an augmentation technique by hallucinating other $M_i$ planning problems $S_i^j$ for which the existing motion plan $P_i$ would also be optimal: $\mathcal{D}^{S*} = \{d_i^*\}_{i=1}^N = \{ \{S_i^j\}_{j=1}^{M_i}, P_i\}_{i=1}^N$. 

However, for multi-goal motion planning, such an assumed mapping from S-spaces alone to motion plans does not hold. A motion planner needs to map from an S-space, start $s_s$, and goal $s_g$ to a motion plan $P$, i.e., $P=f(S, s_s, s_g)$. Thus, we generalize the  previous augmented single-start-goal motion memory dataset $\mathcal{D}^{S*}$ into a multi-start-goal dataset: 
\begin{align}
    \mathcal{D}^{M*} &= \{{}_k\mathcal{D}^{S*}\}_{k=1}^K =\{\{{}_{k}d_i^*\}_{i=1}^{{}_kN}\}_{k=1}^K \nonumber\\
    &= \{ \{ \{{}_kS_i^j\}_{j=1}^{{}_kM_i}, {}_kP_i\}_{i=1}^{{}_kN} \}_{k=1}^K, \nonumber
\end{align}
where $K$ is the number of start-goal pairs, $K = {}_{|\Goals|}\!P_2$ (permutation of selecting and ordering 2 goals from $|\Goals|$ goals). Left subscript $k$ denotes the $k$th start-goal pair. 
Each start and goal are denoted as ${}_ks_s \in \Goals$ and ${}_ks_g \in \Goals$. 

% Multi-Goal Motion Memory also includes an augmentation technique of dataset (\(\mathcal{D}\)) with hallucinated planning problems to enhance the capability to adapt to new, unseen challenges in multi-goal settings. The available dataset generated by solving different multi-goal motion planning problems in the past by a motion planner will be $\mathcal{D}' = \{\mathcal{D}\}_{k=1}^K = \{\{d_i\}_{i=1}^N\}_{k=1}^K$, where $K$ is the number of start-goal pairs, $K = {}^n\!P_2
% $ and n is the number of goals in \Goals. With experience hallucination techniques, the original dataset $\mathcal{D}'$ is augmented to 
% \begin{equation}
%     \mathcal{D}^* = \{\{d_i^*\}_{i=1}^N\}_{k=1}^K = \{ \{ \{S_i^j\}_{j=1}^{M_i}, P_i\}_{i=1}^N\}_{k=1}^K, 
%     \nonumber
% \end{equation}
% where $M_i$ is the number of total generated planning problems for solution $i$, each past planning solution $P_i$ is paired with one original planning problem $S_i$ and a set of $M_i$ generated planning problems $\{S_i^j\}_{j=1}^{M_i}$ for one start-goal pair, for which $P_i$ is optimal.

The augmented dataset ($\mathcal{D}^{M*}$) is used to learn the motion memory so that given a specific start-goal pair, $s_s$ and $s_g$, and a S-space $S$ it will predict a most likely collision-free and near-optimal path $P$. 
Despite many different data-driven approaches to approximate this function, inspired by the success of single-start-goal motion memory, we utilize a similar representation learning technique for each start-goal pair $k$:  An encoder, \(e_{\theta_k}(\cdot)\), parameterized by \({\theta_k}\), maps S-spaces to latent representations by minimizing a triplet loss: 
\begin{equation}
    \begin{split}
        &L_k(\langle {}_kS^a, {}_kS^s, {}_kS^d \rangle) = \max(||e_{\theta_k}({}_kS^a)- \\
&e_{\theta_k}({}_kS^s)|| - ||e_{\theta_k}({}_kS^a) - e_{\theta_k}({}_kS^d)|| + \delta, 0), 
    \end{split}
    \label{eqn::triplet}
    \nonumber
\end{equation}
where ${}_kS^a$ and ${}_kS^s$ are sampled from the same set of ${}_kM_i$ planning problems for start-goal pair $k$ where ${}_kP_i$ is optimal, and sample ${}_kS^d$ sampled from other problem sets where $P_i$ is not optimal. \(\delta\) is a margin that ensures a minimum distance between similar and dissimilar planning problem pairs. This process creates $K$ different representation spaces, each with ${}_kN$ different latent clusters, which correspond to the ${}_kN$ existing motion plans $\{{}_kP_i\}_{i=1}^{{}_kN}$ for each of the $K$ start-goal pairs. The cluster centroids are computed as 
\begin{equation}
    {}_kc_i = \frac{1}{{}_kM_i}\sum_{j=1}^{{}_kM_i} e_{\theta_k}({}_kS_i^j).
    \nonumber
\end{equation}

Upon encountering a new planning problem \({}_kS_{N+1}\) for the $k$th start-goal pair, the latent representation \(l_{N+1} = e_{\theta_k}({}_kS_{N+1})\) is calculated and compared against the centroids of existing clusters (\({}_kc_i\)) in the latent space. The closest cluster indicates the most suitable pre-existing motion plan (\({}_kP_{i^*}\)):
\begin{equation}
    i^* = \argmin_{i} ||e_{\theta_k}({}_kS_{N+1}) - {}_kc_i||. 
    \nonumber
\end{equation}
where \(i^*\) identifies the selected motion plan that closely approximates the optimal solution for \({}_kS_{N+1}\) from start ${}_ks_s \in \Goals$ to goal ${}_ks_g \in \Goals$. 

% After predicting a dynamically-feasible trajectory for each start-goal pair, $\zeta_i: \{0, \ldots, \ell\} \rightarrow S$ where $\zeta_i\in P_{i^*}$, a roadmap $\Phi$ is constructed to find paths to all goals from each other. Instead of sampling from the entire $S_{free}$, we sample along the co-ordinates of $\zeta_i$ and find all goal-to-goal paths by connecting the sampled regions by using  Dijkstra’s algorithm.

% Then we build a motion tree $\Tree$ rooted at $s_s$ which dynamically expands within $S_{free}$. A motion tree \(\Tree\), rooted at \(s_s\), dynamically expands within the $S_{free}$ considering robot dynamics and obstacles. If $\TreeNode \in \Tree$ is a node from the motion tree, targets are sampled around $\zeta$ to expand $\Tree$ from $\TreeNode$ toward a goal $g \in \Goals$.

\subsection{Multi-Goal Motion Planner Integration}
Using the predicted motion plans \({}_kP_{i^*}, \forall k\), dynamically-feasible trajectories $\zeta_k$ can be produced by Eqn.~\eqref{eqn:simulate} for each start-goal pair $k$. 
Then a motion map \(\Phi\) is constructed to facilitate the discovery of paths connecting all goals by sampling along the coordinates of all \(\zeta_k\) within the free space \(S_{free}\), rather than the entire space, and linking these sampled regions using Dijkstra’s algorithm to establish goal-to-goal paths (Algorithm~\ref{alg:multi-goal-motion-planner} line 1). 
% Subsequently, a motion tree, \(\Tree\), with its root at \(s_\Var{init}\), is incrementally developed within \(S_{free}\), taking into account the dynamics of the robot and potential obstacles. In this process, targets around \(\zeta_k\) are sampled to extend \(\Tree\) from any given node, \(\TreeNode \in \Tree\), towards any goal \(g \in \Goals\).

To compute a TSP tour \(\tau\) originating from \(s_\Var{init}\) and targeting the remaining goals, we first evaluate the distance set \(\textrm{DIS} = \{\textrm{dis}_k\}_{k=1}^K\) for each predicted trajectory \(\zeta_k\). Beginning from \(s_\Var{init}\), the tour \(\tau\) is planned by iteratively moving towards the next goal with the lowest \(\textrm{dis}_k\) from the current goal (Algorithm~\ref{alg:multi-goal-motion-planner} line 2).

With the tour $\tau$ determined efficiently, a motion tree, \(\Tree\), with its root at \(s_\Var{init}\), is incrementally developed within \(S_{free}\), taking into account the dynamics of the robot and potential obstacles. In this process, targets around \(\zeta_k\) are sampled to extend \(\Tree\) from any given node, \(\TreeNode \in \Tree\), towards any goal \(g \in \Goals\). The motion tree (\(\Tree\)) is expanded to sequentially reach all goals according to $\tau$, guided by the Multi-Goal Motion Memory. This involves prioritizing states near the predicted optimal trajectories \(\zeta_k\), thereby focusing the exploration on promising areas of \(S_{free}\) and enhancing planning efficiency (Algorithm~\ref{alg:multi-goal-motion-planner} line 3). 

% \subsection{Motion Memory for Single-Goal Motion Planning}
% \label{sec:MotionMemory}

% When a group $\Group_\Pair{p, \Var{g}}$ is first created, we invoke a TSP solver to determine a tour in which to visit the remaining goals $\Goals\setminus \Var{goals}$ starting from $p$ (Alg.~\ref{alg:Partition}:12), i.e.,
% \begin{equation}
%  \Group_\Pair{p, \Var{g}}.\Var{tour} \gets \Function{TSPsolver}(\CostMatrix)
% \end{equation}

% An entry $\CostMatrix_{i,j}$ in the cost matrix corresponds to a single-goal motion-planning problem (from $p$ to a goal or from a goal to another goal). We can use the cost from the roadmap or the predicted distance provided by Motion Memory (Alg.~\ref{alg:Partition}:11), i.e., 

% \begin{equation}
% \CostMatrix_{i, j} \gets \Function{RMCost}(i, j).
% \end{equation}

% \subsection{Training on Single-Start-Goal Motion Memory}

\section{Implementation}

\begin{figure}[t]
    \centering 
    \includegraphics[width=0.95\columnwidth]{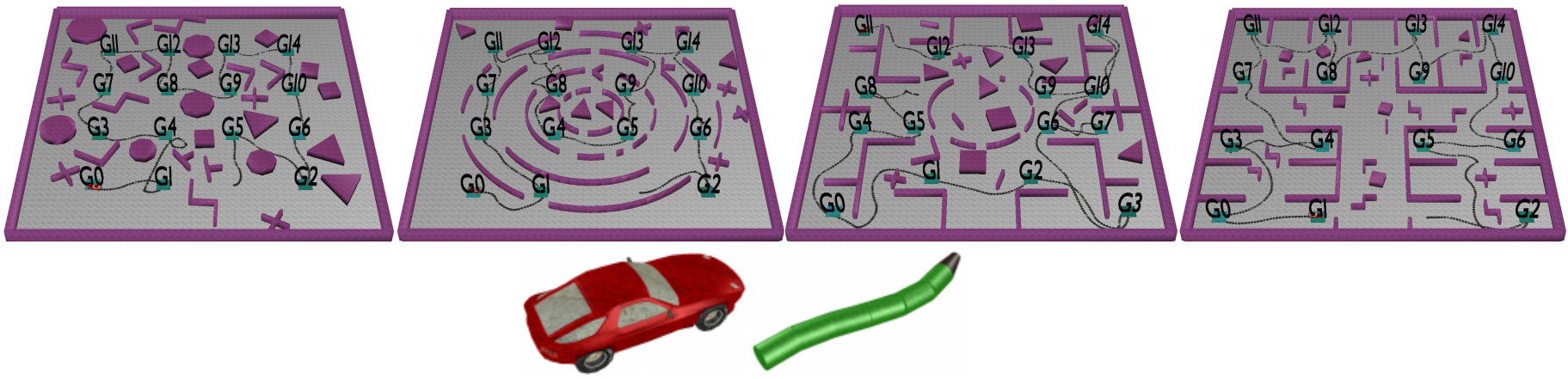} 
    \caption{Different Planning Problem Classes: \texttt{random}, \texttt{curve}, \texttt{maze}, and \texttt{storage} (Left to Right) and Car and Snake Robot.}
    \label{fig::env}
\end{figure}

We implement Multi-Goal Motion Memory with a multi-goal motion planner, which follows the same structure defined in Sec.~\ref{sec:preliminaries}, to showcase the improved planning time without sacrificing plan quality. 

\subsection{Guided Motion Map Construction}
\label{sec:Roadmap}

To build a motion map guided by Multi-Goal Motion Memory, the approach first generates all trajectories for all start-goal and goal-goal pairs, as shown in Alg.~\ref{alg:Main}a. This process is efficient as it relies on retrieving the most suitable trajectories from previous experiences. Since the current environment is possibly different from environments used during training, there is some possibility that parts of the retrieved trajectories could collide with the obstacles. For this reason, a trajectory repair process is used which samples collision-free states around all the retrieved trajectories. These collision-free states then serve as nodes for our motion map. The next step is to connect neighboring states, discarding those edges that are in collision, giving rise to a motion map that captures the connectivity of the environment. Dijkstra's algorithm is then used to find shortest paths for each goal pair.  If the motion map does not contain the path between a certain goal pair, we call the motion memory again to use the predicted trajectory as an alternative path. This process is repeated until all goal pairs are connected by paths. Pseudocode can be found in Alg.~\ref{alg:Main}b.

\begin{algorithm}[!t]
\caption{Multi-Goal Motion Memory}
\label{alg:Main}

% \begin{algorithmic}[1]
% \STATE $\Tree \gets \Function{CreateMotionTree}(s_\Var{init})$
% \STATE $\Lambda, \CostMatrix \gets \Function{CallMMemory}(s_\Var{init}, \mathcal{G})$
% \STATE $\Function{GenerateMotionMap}(s_\Var{init}, \Goals, \Lambda)$

% \STATE $\Group \gets \emptyset$; $\Function{UpdateGroups}(\Group, \TreeNode_\Var{init},\Lambda , \CostMatrixAll)$
% \WHILE{$\Function{time} < t_\Var{max}$}
% \STATE $\Group_{p, \Var{g}} \gets \Function{SelectGroup}(\Group)$
% \STATE $\TreeNode \gets \Function{SelectNode}(\Group_{p, \Var{g}}.\Var{nodes})$
% \STATE $g \gets$ first goal in $\Group_{p, \Var{g}}.\Var{tour}$
% \STATE $p_\Var{target} \gets \Function{SelectTarget}(\TreeNode)$
% \STATE $\Function{extend}(\Tree, \TreeNode, p_\Var{target}, \Group)$
% \FOR{each new node $\TreeNode_\Var{new}$ added by $\Function{extend}$}
% \STATE \textbf{if} $\TreeNode_\Var{new}.\Var{g} = \Goals$ \textbf{return} $\Traj_\Tree(\TreeNode_\Var{new})$
% \STATE $\Function{UpdateGroups}(\Group, \TreeNode_\Var{new})$
% \ENDFOR
% \ENDWHILE

% \end{algorithmic}

\textbf{Input}: goals $\mathcal{G}$; a set of coordinates for the predicted trajectory $\Lambda$; length of predicted trajectories $\CostMatrix$
\vspace*{1mm}
\hrule

\hrule
\vspace*{1mm}
a) $\Function{CallMMemory}(s_\Var{init}, \Goals)$
\label{alg:MMemory}

\begin{algorithmic}[1]
\STATE $\Lambda \gets \Null; \CostMatrix \gets \Null$

\STATE \textbf{for} $i \in \Goals$ \textbf{do} $\Lambda_{s_\Var{init}, i}, \CostMatrix_{s_\Var{init}, i} \gets \Function{MMemory}(s_\Var{init}, i)$
\STATE \textbf{for} $(i, j) \in \Goals \times \Goals$ \textbf{do}
$\Lambda_{i, j}, \CostMatrix_{i, j} \gets \Function{MMemory}(i, j)$

\end{algorithmic} 

\hrule
\vspace*{1mm}
b) $\Function{GenerateMotionMap}(s_\Var{init}, \Goals, \Lambda)$
\label{alg:generatemap}

\begin{algorithmic}[1]

\FOR{$i \in \Goals$}
\FOR{$p \in \Lambda_{s_\Var{init}, i}$}
\STATE $p_\Var{near} \gets \Function{SampleInVicinity}(p)$
\IF{$\Function{CheckCollision}(p_\Var{near})$}
\STATE
$\Function{AddMotionMap}(p_\Var{near})$
\ENDIF
\ENDFOR
\ENDFOR

\FOR{$(i, j) \in \Goals \times \Goals$}
\FOR{$p \in \Lambda_{i, j}$}
\STATE $p_\Var{near} \gets \Function{SampleInVicinity}(p)$
\IF{$\Function{CheckCollision}(p_\Var{near})$}
\STATE
$\Function{AddMotionMap}(p_\Var{near})$
\ENDIF
\ENDFOR
\ENDFOR

\STATE $\Function{GeneratePaths}(\Goals, \Goals)$
\FOR{$(i, j) \in \Goals \times \Goals$}
\STATE \textbf{if} $\neg \Function{FindPath}(i, j)$ \textbf{then} $\Var{path}_{i, j} \gets \Lambda_{i, j}$ 
\ENDFOR

\end{algorithmic} 

\hrule
\vspace*{1mm}
c) $\Function{SelectTarget}(\TreeNode, \Lambda)$
\label{alg:SelectTarget}
\begin{algorithmic}[1]

\STATE $p_\Var{target} \gets \Null$; $g \gets \Function{FirstGoalFromTour}(\TreeNode.\Var{tour})$
\FOR{several times}

\IF{$\Lambda_{\TreeNode.s, g} = \Null$}
\STATE \textbf{return} $p_\Var{target} \gets \Function{RandomSample}()$
\ENDIF
\STATE $p \gets \Function{SelectFirstCoordinate}(\Lambda_{\TreeNode.s, g})$
\STATE $p_\Var{target} \gets \Function{GenerateSampleInVicinity}(p)$
\IF{$\Function{CollisionCheck}(\TreeNode.s, p_\Var{target})$}
\STATE $\Function{RemoveFirstCoordinate}(\Lambda_{\TreeNode.s, g})$
\STATE \textbf{return}  $p_\Var{target}$
\ENDIF
\ENDFOR
\STATE \textbf{return} $p_\Var{target} \gets \Function{RandomSample}()$

\end{algorithmic}
\end{algorithm}

\subsection{Multi-Goal Motion Planning Augmented with Motion Memory}
\label{sec:Overall}

The overall approach starts by rooting the motion tree $\Tree$ at the initial state. The first stage (Alg.~\ref{alg:Main}(a)) leverages the motion-memory framework to predict paths for each start-goal and goal-goal pair. The predicted paths are stored in a set $\Lambda$ and their distances are stored in a cost matrix $\CostMatrix$. The second stage (Alg.~\ref{alg:Main}(b)), as described in Section~\ref{sec:Roadmap}, builds the motion map by utilizing the predicted paths $\Lambda$, repairing via sampling when necessary, in order to connect each start-goal and goal-goal pair with a collision-free path.

The third stage utilizes the motion map, the cost matrix $\CostMatrix$, and a TSP solver to effectively guide the motion-tree expansion. The motion-tree is partitioned into groups, denoted by $\Group$, based on the goals and nodes in the motion map that have been reached. New groups are created when new goals or new nodes in the motion map are reached. 
More details about the group partition can be found in \cite{plaku2018multi, lu2023leveraging}. 

When a new group is created, the TSP solver is invoked to compute an optimal tour of how to visit the remaining goals associated with that group. The TSP solver relies on the cost matrix $\CostMatrix$, which stores the set of distances produced by the motion memory predictions. The cost of the tour is used to define the group's priority, giving higher priority to groups associated with low-cost tours.

Starting from the initial state, the motion-tree $\Tree$ is expanded incrementally. During each iteration, the group with the highest priority is selected. A node $\TreeNode$ then is selected at random from the group and attempts are then made to extend the motion tree from $\TreeNode$ along the path to the first goal in the group's tour. Specifically, the objective is to reach the points along the path in succession. For this reason, as described in Alg.~\ref{alg:Main}c,  the first point from the path is retrieved and the target is generated near it (by sampling at random inside a small circle). If the target is in collision, the sampling is repeated again. If no collision-free target is generated after several attempts, then the target is generated by sampling a collision-free point at random from the entire space.

Once the target is generated, a PID controller is used to extend the motion tree from the selected $\TreeNode$ toward the target. If the target is reached, the first point is removed from the path (so that the expansion can continue toward the next point in the path). Otherwise, this process is repeated several times. If after several times, the target is not reached, the group's priority is lowered and the expansion from this group is abandoned. In this way, the approach proceeds with a new iteration, possibly selecting a different group for expansion. As demonstrated by the experiments in the next section, this results in an effective way to expand the motion-tree along tours
that leverage previous planning experiences, as generated by the motion memory framework.

\section{Experiments and Results}
\label{sec:ExpResults}

\begin{figure}
\centering
\includegraphics[width=0.97\columnwidth]{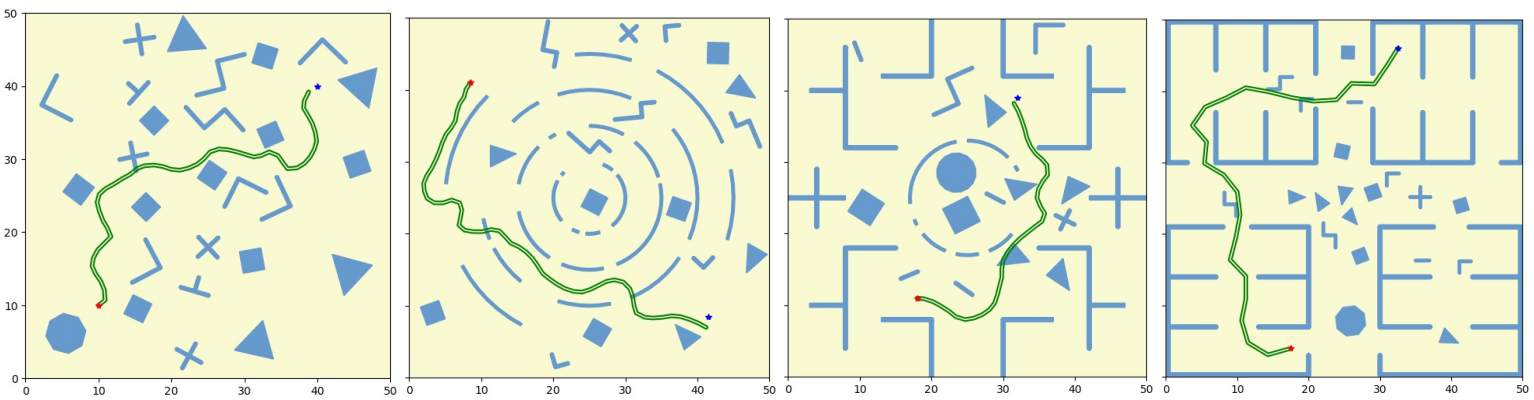}
\\[-1mm]
\caption{Example Goal-to-Goal Path Predictions by Multi-Goal Motion Memory in \texttt{random}, \texttt{curve}, \texttt{maze}, and \texttt{storage} (Left to Right).}
\label{fig:visu}
\end{figure}

\begin{figure*}
  \centering  
  \includegraphics[width=\textwidth]{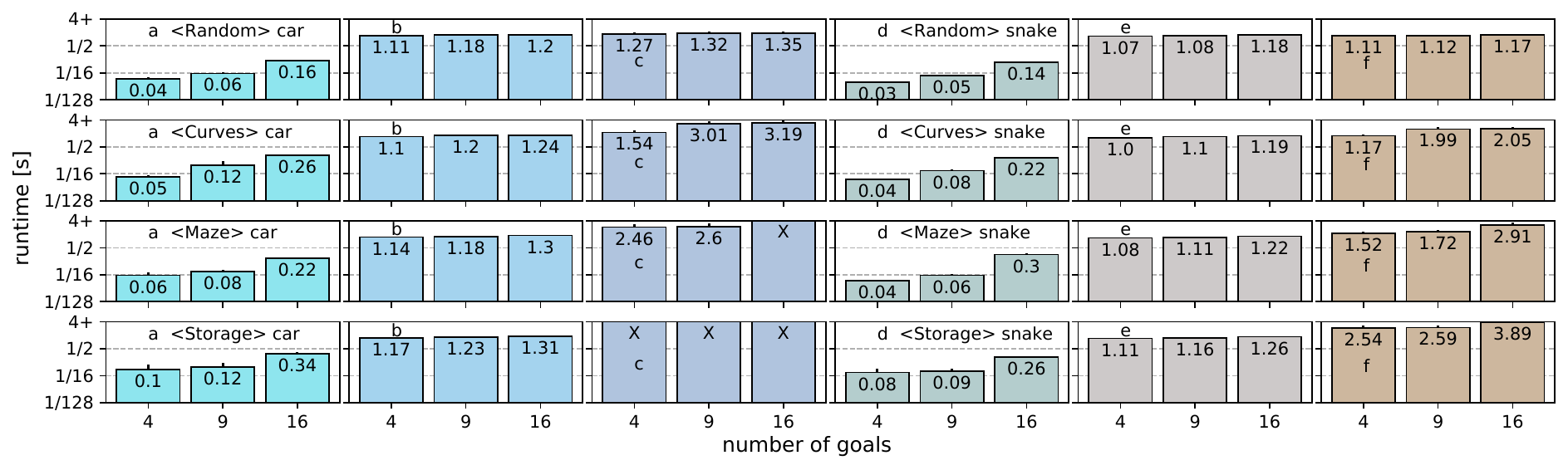}\\[-2mm]
  \caption{Runtime Results with Different Numbers of Goals: (a) Multi-Goal Motion Memory [car]; (b) DROMOS [car]; (c) SequentialRRT [car]; (d) Multi-Goal Motion Memory [snake]; (e) DROMOS [snake]; and (f) SequentialRRT [snake]. Entries marked with $X$ indicate failure.}
  \label{fig:ResRuntimeGoals}
\end{figure*}

\begin{figure*}
  \centering  
\includegraphics[width=2.0\columnwidth]{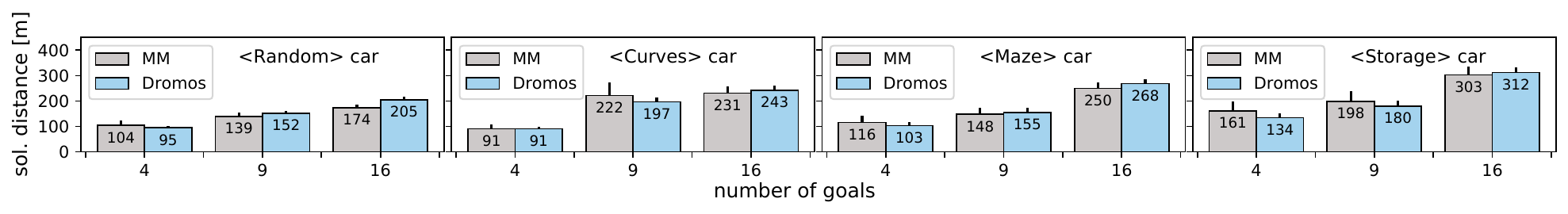}\\[-1mm]
  \includegraphics[width=2.0\columnwidth]{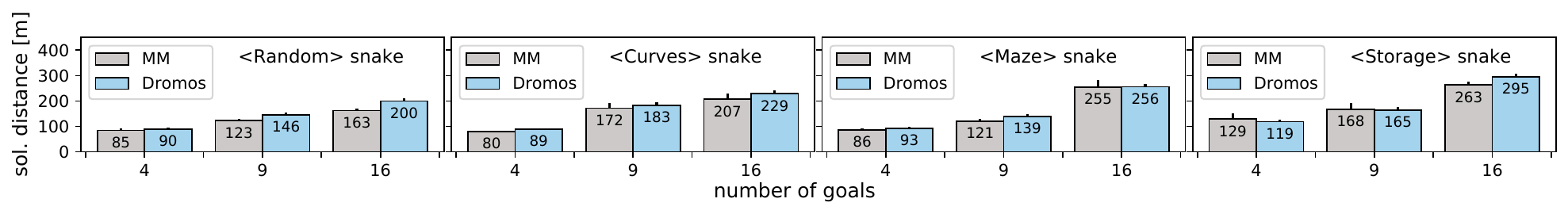}\\[-2mm]
   \caption{Solution Distance with Different number of Goals: (a) Multi-Goal Motion Memory (MM); (b) DROMOS.}
  \label{fig:ResDistanceGoals}
\end{figure*}

Experiments are conducted using a vehicle and snake-like robot operating in unstructured, obstacle-rich environments. The scenes are shown in Figs.~\ref{fig:Main} and \ref{fig::env}, while the robot model is described in Section~\ref{sec:Problem}. Fig.~\ref{fig:Main} left depicts the area a classical motion planners have to explore when facing a new environment, while Fig.~\ref{fig:Main} right shows the region explored using Multi-Goal Motion Memory, revealing a significant reduction in the resources required for planning. 
% These represent challenging motion-planning problems where the robot has to carefully move around obstacles and pass through narrow passages to reach the goals.

\subsection{Experimental Setup}
\subsubsection{Methods for Comparisons}
Multi-Goal Motion Memory (Alg.~\ref{alg:Main}) is compared to DROMOS~\cite{plaku2018multi}, a state-of-the-art motion planner specifically designed for multi-goal problems. DROMOS was shown to be significantly faster than other planners. DROMOS uses a roadmap, shortest roadmap paths, and a TSP solver to guide the motion-tree expansion. As a baseline, we also used a sequential version of RRT~\cite{lavalle2001randomized}, denoted by SequentialRRT, which aims to reach the goals one at a time, by going first to the nearest goal.

% The runtime limit for each method is $4$s per run.

\subsubsection{Multi-Goal Problem Instances and Measuring Performance}
Our experiments include four scenes: \texttt{random}, \texttt{curve}, \texttt{maze}, and \texttt{storage}. For each scene, the experiments are conducted with 4 ($2\times2$), 9 ($3\times3$), and 16 ($4\times4$) goals. These goal locations are fixed and the robot starts from one randomly-chosen goal. For a given scene and number of goals, we generate 400 problem instances. Each method is run on each of the instances. An upper bound of $10$s is set for each run. When calculating the performance statistics, we remove the lower and upper 25\% instances in terms of runtime to avoid the influence of outliers. The runtime includes everything from reading the input files to reporting that a solution is found. The solution length is measured as the distance traveled by the robot along the solution trajectory.

% \begin{figure*}
%   \centering
%   \includegraphics[width=2.0\columnwidth]{figures/ML1.pdf}\\[-1mm]
%    \caption{Ablation Studies on Different Probability of Selection and Number of Neighbors for Each Group Values.}
%   \label{fig:ResParams}
% \end{figure*}

\subsubsection{Training Datasets For Multi-Goal Motion Memory}
% We regard the starting position as a goal, resulting in 2x2, 3x3, and 4x4 goals for the cases with 3, 8, and 15 goals respectively. 
For each different scene and each pair of goals, we generate 100 original planning problems and subsequently 5000 augmented environments for these different problems by slightly rearranging obstacles close to the motion plan and randomly shuffling obstacles in other places. These images are represented as 2D grids and input into convolutional neural networks for training. Each model predicts path for different goal pairs in a new set of 100 test problems. Four example predictions are shown in Fig.~\ref{fig:visu}. Despite some collisions, the prediction can guide motion planners to efficiently find a path. In our experiments, the motion memory model trained for the car is directly used for the snake-like robot.
% The concept of top-5 is to compare the results of the model with the best 5 IDs given by the model to predict the accuracy of its model.

% \subsubsection{Single-Goal Motion Memory Visualization}
% After we run 400 problem instances, we randomly select an instance in each environment and randomly select a starting position and a goal position. We visualize a trajectory predicted by the motion memory, as shown in Fig.~\ref{fig:visu}. The results show that although the trajectories predicted by the single-goal motion memory have some collisions with obstacles, overall they appear quite feasible.

\subsubsection{Computing Resources}
The experiments are run on an AMD Ryzen 9 5900X (3.7 GHz CPU) using Ubuntu 20.04. The motion planning code is written in C++ and compiled using g++-9.4.0, while the code for training the models is written in Python 3.8.

\subsection{Results}

\subsubsection{Runtime Results with Different Number of Goals} 
Fig.~\ref{fig:ResRuntimeGoals} shows the runtime results for $4$, $9$ and $16$ goals. The results indicate that Multi-Goal Motion Memory is significantly faster than DROMOS and SequentialRRT.  

The predicted trajectories can effectively guide motionmap construction, estimate goal-to-goal distance in the TSP solver, and bias the expansion of the motion tree along low-cost tours. SequentialRRT is the least effective among the three approaches. It lacks guidance and often wastes many resources during the motion-tree expansion process due to its randomized characteristics. DROMOS can solve multi-goal motion planning problems by relying on roadmaps to guide the motion-tree expansion, but it has difficulty in challenging environments since the roadmap paths do not necessarily account for the robot dynamics.

Parts d, e, and f in Fig.~\ref{fig:ResRuntimeGoals} show the snake-like results produced by these three approaches. Experimental results also show that our motion memory method has a significant acceleration effect even on a different robot.

\subsubsection{Solution Distance Results} Fig.~\ref{fig:ResDistanceGoals} shows the solution distances, i.e., plan quality. SequentialRRT is not inlcuded due to its frequent failure. The results show that Multi-Goal Motion Memory produces similar solution distances as DROMOS, and even outperforms DROMOS in many environments. Such results highlight that Multi-Goal Motion Memory significantly improves multi-goal motion planning efficiency, without sacrificing plan quality. 

% \subsubsection{Impact of Various Parameters} Fig.~\ref{fig:ResParams} shows two ablation studies. The first experiment corresponds to the problem mentioned in~\ref{sec:Roadmap}: once the number of configurations exceeds the threshold, we select the configurations generated later with a probability of 20\%, 50\% and 80\%. The second experiment corresponds to Sec.~\ref{sec:Partition}: each group has a central node and its neighbor nodes while the maximum number of neighbor nodes that the group can store is varied among 5, 10 and 15. Overall, the best results can be achieved using the default values, i.e., $\alpha > 0.5$ and $10$ neighbors for each group. 

\section{Conclusions and Future Work}
\label{sec:Conclusions}

Leveraging past planning experiences, Multi-Goal Motion Memory is an efficient approach to augment multi-goal motion planners  to solve challenging problems in unstructured, obstacle-rich environments while considering robot dynamics. The approach's scalability is demonstrated by successfully handling a variety of numbers of goals with a large number of changing environments. 
% A crucial aspect of the approach was to leverage machine-learning models trained on single-goal motion-planning problems to guide the motion-tree expansion.

This work suggests several avenues for future research. Enhancing the machine-learning models may further improve the overall performance and reduce extra computation overhead. We demonstrate that machine-learning models trained on one robot model has the potential to transfer to other robot models. Therefore, future research can focus on generalizing training across similar and different robot models. Additionally, another interesting direction is to explore extending Multi-Goal Motion Memory to multiple robots.

\IEEEpeerreviewmaketitle

\bibliographystyle{IEEEtran}
\bibliography{ref.bib}

\end{document}